\documentclass[conference]{IEEEtran}
\IEEEoverridecommandlockouts
\usepackage{cite}
\usepackage{amsmath,amssymb,amsfonts}
\usepackage{algorithmic}
\usepackage{rotating}
\usepackage{graphicx}
\usepackage{textcomp}
\usepackage[algo2e, ruled]{algorithm2e}
\usepackage{xcolor}
\usepackage{soul,flushend}
\def\BibTeX{{\rm B\kern-.05em{\sc i\kern-.025em b}\kern-.08em
    T\kern-.1667em\lower.7ex\hbox{E}\kern-.125emX}}

\graphicspath{ {./images/} }

\usepackage{tikz}
\usepackage{textcomp}
\usepackage{hyperref}
\usepackage{lipsum}

\newcommand\copyrighttext{%
  \footnotesize \textcopyright 2023 IEEE. Personal use of this material is permitted.
  Permission from IEEE must be obtained for all other uses, in any current or future
  media, including reprinting/republishing this material for advertising or promotional
  purposes, creating new collective works, for resale or redistribution to servers or
  lists, or reuse of any copyrighted component of this work in other works.
  DOI: \href{https://ieeexplore.ieee.org/abstract/document/10371958}{10.1109/SSCI52147.2023.10371958}}
\newcommand\copyrightnotice{%
\begin{tikzpicture}[remember picture,overlay]
\node[anchor=south,yshift=10pt] at (current page.south) {\fbox{\parbox{\dimexpr\textwidth-\fboxsep-\fboxrule\relax}{\copyrighttext}}};
\end{tikzpicture}%
}

\begin{document}
\title{Training Artificial Neural Networks by \\ Coordinate Search Algorithm}

\author{
\IEEEauthorblockN{Ehsan Rokhsatyazdi$^{*1}$, Shahryar Rahnamayan$^{*2}$, SMIEEE, Sevil Zanjani Miyandoab$^{*1}$,\\ Azam Asilian Bidgoli$^{*3}$, H.R. Tizhoosh$^{4}$, SMIEEE}\\
\IEEEauthorblockA{
\textit{*Nature-Inspired Computational Intelligence (NICI) Lab} \\
\textit{$^1$Department of Electrical, Computer, and Software Engineering, Ontario Tech University, Oshawa, 
ON, Canada}\\
\textit{$^2$Department of Engineering, Brock University, St. Catharines, ON, Canada}\\
\textit{$^3$Faculty of Science, Wilfrid Laurier University, Waterloo, ON, Canada}\\
\textit{$^4$Rhazes Lab, Department of Artificial Intelligence and Informatics, Mayo Clinic, Rochester, MN, USA}\\
}}

\maketitle
\copyrightnotice

\begin{abstract}
Training Artificial Neural Networks (ANNs) poses a challenging and critical problem in machine learning. Despite the effectiveness of gradient-based learning methods, such as Stochastic Gradient Descent (SGD), in training neural networks, they do have several limitations. For instance, they require differentiable activation functions, and cannot optimize a model based on several independent non-differentiable loss functions simultaneously; for example, the F1-score, which is used during testing, can be used during training when a gradient-free optimization algorithm is utilized. Furthermore, the training (i.e., optimization of weights) in any DNN can be possible with a small size of the training dataset. To address these concerns, we propose an efficient version of the gradient-free Coordinate Search (CS) algorithm, an instance of General Pattern Search (GPS) methods, for training (i.e., optimizing) neural networks. The proposed algorithm can be used with non-differentiable activation functions and tailored to multi-objective/multi-loss problems. Finding the optimal values for weights of ANNs is a large-scale optimization problem. Therefore, instead of finding the optimal value for each variable, which is the common technique in classical CS, we accelerate optimization and convergence by bundling the variables (i.e., weights). In fact, this strategy is a form of dimension reduction for optimization problems. Based on the experimental results, the proposed method is comparable with the SGD algorithm, and in some cases, it outperforms the gradient-based approach. Particularly, in situations with insufficient labeled training data, the proposed CS method performs better. The performance plots demonstrate a high convergence rate, highlighting the capability of our suggested method to find a reasonable solution with fewer function calls. As of now, the only practical and efficient way of training ANNs with hundreds of thousands of weights is gradient-based algorithms such as SGD or Adam. In this paper we introduce an alternative method for training ANN.

\end{abstract}


\section{Introduction}

Optimizing the weights in an Artificial Neural Network (ANN), also called \emph{training}, is one of the most significant and challenging machine learning problems\cite{kaveh2022application} and is still an open research direction. Since there are thousands or millions of weights in the state-of-the-art neural networks, it is considered a huge-scale optimization problem, and solving such high-dimensional problems is very expensive in terms of time and memory complexities. Metaheuristic methods are one way to train ANNs.

Kaveh et al. \cite{kaveh2022application} reviewed recent developments in metaheuristic algorithms for deep learning and training ANNs. They have compared these algorithms based on exploitation and exploration abilities, convergence speed, finding the global optimum, hyper-parameter setting, and implementation. According to their results, hybrid metaheuristic-convolutional neural network architectures perform better than other methods in many medical image classification applications and are effective in medical applications.


Metaheuristic optimizers are also useful when back-propagation-based optimization methods face crucial limitations and do not perform properly. For example, in Spiking Neural Networks (SNNs) \cite{javanshir2023training}, the nature of the spiking neurons is discontinuous and non-differentiable. Accordingly, Javanshir et al. have proposed a novel metaheuristic-based optimizer for SNNs, and shown that it outperforms other methods, e.g., genetic algorithm (GA), differential evolution (DE), particle swarm optimization (PSO), and harmony search (HS).

Several metaheuristic-based optimization algorithms have been introduced for training ANNs and shown significant results. Improved GA (IGA) \cite{leung2003tuning}, improved PSO \cite{meissner2006optimized}, PSO-MLFFNN \cite{geethanjali2008pso}, and ESPNet \cite{yu2008evolving} are some examples. Generally speaking, metaheuristic or swarm optimization algorithms can be used for training, but they are usually very time-consuming, or used for very small size networks with a small number of weights \cite{zhou2006pso, carvalho2007particle, grimaldi2004pso, ilonen2003differential}. Hence, gradient-based training methods are often preferred, since they are faster and more efficient. These optimization algorithms work by relying on derivatives (for one-dimensional functions), and gradients (for more than one-dimensional problems) \cite{jameson1995gradient}; therefore, activation functions must be differentiable when using these methods. Moreover, despite the efficiency of the existing state-of-the-art gradient-based methods, e.g., Adam \cite{kingma2014adam}, they cannot be employed in the case of multiple loss/objective functions (multi-objective problems). 

On the other hand, Coordinate Descent (CD) algorithms try to tackle optimization problems by solving a sequence of simple optimization problems~\cite{schwefel1993evolution}. Coordinate Search (CS) is also a Generalized Pattern Search (GPS) algorithm. These algorithms are derivative-free and the objective function is calculated at a specific number of sample points along a suitable set of search directions in each iteration \cite{bogani2009generalized, tzinis2019bootstrapped}. Hence, they are a category of decomposition-based algorithms. At each iteration, CD optimizes one or a block of coordinates (variables) while fixing all other coordinates or blocks~\cite{bidgoli2021memetic}. In fact, the idea behind the CD algorithm is that, when directional derivatives are unavailable or difficult to compute, one-dimensional minimization can be replaced to approximate an acceptable solution~\cite{frandi2014coordinate}. In numerical linear algebra or arithmetic optimization, gradient information is required to apply CD. However, when exact gradient information is not applicable, the method is called coordinate search (CS). In this case, the algorithm benefits function value samplings on coordinates in turn to find a suitable value for each variable. CS can solve black-box optimization problems.


Despite the simplicity of CS and inexact derivative-free minimization, the algorithm still produces acceptable practical results and performs better than many other algorithms in solving expensive optimization problems. Depending on the number of coordinates that are optimized simultaneously, CS can be viewed as two variants. At each iteration, the algorithm may update only one coordinate based on the fitness evaluation of the sampled points while all other coordinates are fixed. In order to accelerate the optimization process, particularly in large- or huge-scale optimization problems, a block-CS can be utilized in which instead of only one coordinate, a block of variables can be updated simultaneously, and consequently, the number of fitness evaluations is decreased dramatically~\cite{tseng2001convergence}.



The  motivation of this study is to develop a simple, general-purpose, and gradient-free ANN optimization (i.e., training) algorithm. Considering that CS is an efficient and popular method for optimizing large-scale problems, we use and tailor it to our purpose. In this research, we show that the proposed CS algorithm can be exploited for training a fully connected neural network with a personal computer using only one CPU in a reasonable time, which is not feasible with other gradient-free algorithms. In addition, in contrast with gradient-based methods, CS can work with non-differentiable activation functions, train an ANN for multi-objective problems with multiple independent loss functions as in \cite{nikbakht2023multi}, and be applied to any ANN/DNN architecture. In other words, it is structure-independent and can be used in a graph structure DNN, e.g., a real brain structure.
Our proposed algorithm shows better performance than gradient-based algorithms in some cases, especially when the training dataset is small, as well. 

This paper is organized as follows: Section II outlines our proposed method, highlighting the steps and employed techniques, Section III provides information regarding experimental settings and the results and analysis derived from our study, and Section IV concludes with remarks.

\section{Proposed Method}
In this section, the components of the proposed method are explained in detail. These techniques include the proposed CS scheme, bundling weights, initialization, and data feeding.

\subsection{Proposed Two-extreme-point CS}
\begin{figure}
\centering
\includegraphics[width=0.90\linewidth]{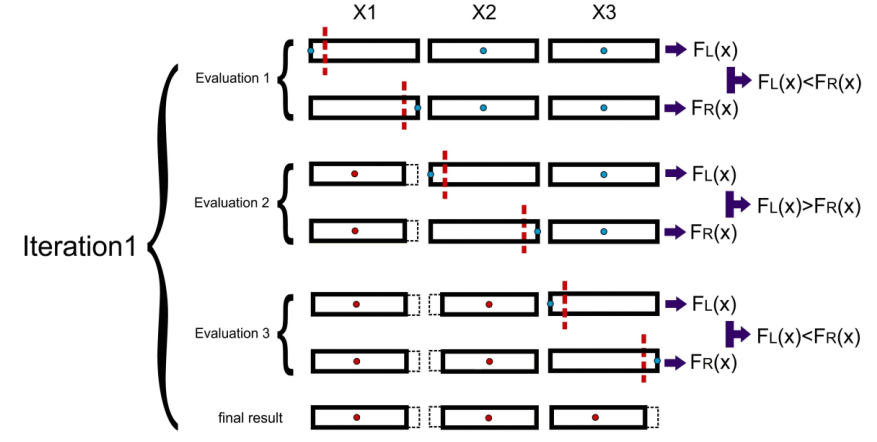}
\caption{An illustration of one iteration in two-extreme-point CS with three variables - For each variable, two extreme points of the box-constraints are evaluated, and the box-constraints are updated based on the best result.}
\label{cs-fig}
\end{figure}

In CS, one variable or block of the variables changes one-by-one in each iteration during the optimization process, and other variables and blocks are kept fixed just like in CD. In two-extreme-point CS, two sample points (i.e., extreme points) for each variable are evaluated in each iteration. Because of its superiority according to our experiments, we prefer this scheme to classical (two-center-point) CS, in which center points of the right and left halves are evaluated for updating the box-constraints, and three-point CS, in which the left, center, and right points are compared. In Algorithm~\ref{alg-cs}, the pseudo-code of the proposed algorithm is provided. At the first iteration, the lower and upper (extreme) values of each variable's box-constraint are selected (i.e., sampled) for function evaluation, while for the next variables, the center points of their box-constraints are considered as the initial values. Fig.~\ref{cs-fig} illustrates this process. In this figure, there are three variables, and the goal is to minimize the fitness function $f(X)$. After comparing the samples of extreme points of each variable, the bounds of box-constraints are updated based on a constraint shrinkage factor (BSF) to shrink the search space. BSF is set to 5\% in our tests. As shown in Algorithm~\ref{alg-cs}, for example, if the lowest point of a box-constraint is a better solution than its uppest point, the uppest 5\% of the interval would be cut and removed. As a result, the interval shrinks toward the better side. Then, the corresponding variable is set to the center of the shrunk region. This process repeats for all variables and updates their values.
This process continues in the second iteration with updated variables and shrunk intervals resulting from the previous iteration. The optimization stops at a predefined number of iterations. 
In this method, the box-constraint for each variable shrinks exponentially with the iteration number.

\begin{algorithm2e}
\SetAlgoLined
\SetKwInOut{Input}{input}\SetKwInOut{Output}{output}
 \Input{ $D$: Dimension of the problem, \\$U$: Upper bounds, $L$: Lower bounds, \\$N_{ite}$: Number of iterations, \\$BSF$: Box-constraint shrinkage factor}
 \Output{ $S$: The best solution found so far }
 \BlankLine

\For{$i\leftarrow 1$ \KwTo $D$}
    {$C_i=\frac{U_i - L_i}{2}$\;}
\For{$j\leftarrow 1$ \KwTo $N_{ite}$}{
\For{$i\leftarrow 1$ \KwTo $D$}
    {
    $S_{iL} = f([C_1, C_2, ..., L_i, ..., C_D])$\;
    $S_{iR} = f([C_1, C_2, ..., U_i, ..., C_D])$\;
    \If {$S_{iL} < S_{iR}$}
          	{$U_i = U_i - (U_i - L_i)\times BSF$\;
           $C_i = \frac{U_i - L_i}{2}$\;
           $S = S_{iL}$\;}
    \If {$S_{iL} > S_{iR}$}
           {$L_i = L_i + (U_i - L_i)\times BSF$\;
           $C_i = \frac{U_i - L_i}{2}$\;
           $S = S_{iR}$\;}
    }
}
\BlankLine
\caption{Two-extreme-point CS}\label{alg-cs}
\end{algorithm2e}

\subsection{Bundling Weights}
In order to accelerate the training procedure, as a large-scale optimization problem, we bundle all variables into $n$ bundles, where $n=N_w / BS$, $N_w$ is the number of variables (weights), and $BS$ stands for the bundle size. In each iteration, all variables in a bundle are optimized simultaneously. Therefore, CS selects the extreme right or left point in their box-constraints for all variables in the bundle based on the fitness evaluation. 
At the beginning of each iteration, variables are shuffled and bundled again to alleviate the dependency of optimization on the order of variables. Nevertheless, the box-constraint differs for each variable in a new bundle, we choose the upper and lower bounds of the box-constraint of each variable as the extreme points.


\subsection{Initialization}
For training a neural network, initialization of the weights is very important. There are different approaches to this issue. The first approach is center initialization which is a wise choice for many applications, and more importantly for high-dimensional problems \cite{rahnamayan2009toward}. But in this special case, the center point of the box-constraint would be zero, and if all variables (except one bundle) are set to zero for initialization, CS can hardly move toward the optimum point. By changing one bundle while other bundles are set to zero, it is more likely to face multiplication by zero, which results in zero for the network output. To prevent this problem, we can choose a large value for bundle sizes to change many weights simultaneously. In this way, inputs can find a non-zero pass to the network output, however, as we will explain later, accuracy may degrade by choosing a large bundle size.

Another way to initialize weights is random initialization, uniform or normal. By normal distribution random initialization, with the mean being zero, we can benefit from center point initialization and avoid plenty of multiplication by zero\cite{rahnamayan2009toward, mahdavi2016center, rahnamayan2009center}.

\subsection{Data Feeding}
\begin{figure}
\centering
\includegraphics[width=0.50\linewidth]{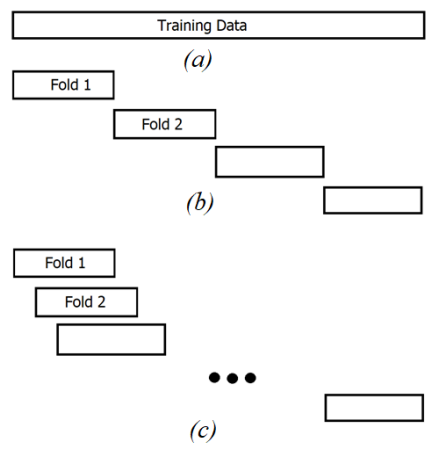}
\caption{Three schemes of feeding the training data into network (a) Feeding whole data as one batch, (b) Separate folds, (c) Sliding fold}
\label{feed}
\end{figure}

Feeding whole training data at once for optimization improves accuracy, but is not the most appropriate choice in terms of efficiency. Instead of using the whole data for training, data can be separated into different folds. For each fitness call to calculate the error value, a fold can be selected for training, which is very similar to batch training \cite{masters2018revisiting}. In this way, the network will see all the training data in multiple phases, and accuracy will not suffer very much. The network trains faster when fed one fold at a time. For example, if the training data is divided into six folds, the training phase can be accelerated by a factor of six. Feeding the network with different folds can be done in two ways. The first one is dividing the training data into several separate folds. The other way is starting with a fold, and sliding that over the data. Fig.~\ref{feed} illustrates three different ways of feeding data. 

\section{Experiments}
\subsection{Network Architecture}
\begin{figure}
\centering
\includegraphics[width=0.70\linewidth]{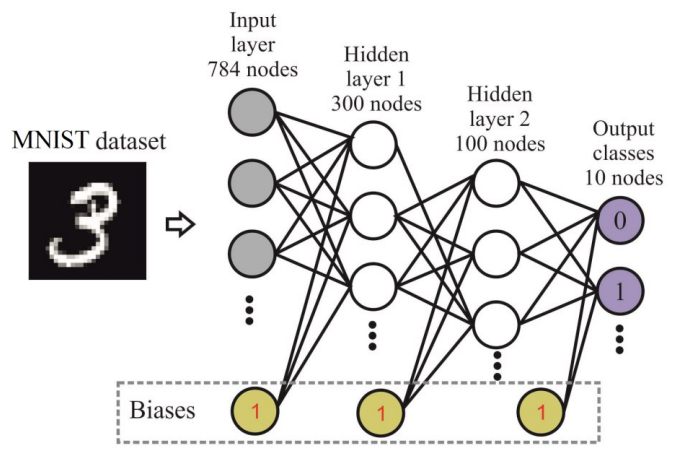}
\caption{Structure and number of the nodes in the fully connected network, used as our framework}
\label{arch}
\end{figure}

Almost in all structures of neural networks, one or more fully connected layers are used. Hence, we have selected a fully connected neural network with two hidden layers as the framework, which is similar to \cite{lu2019not}. The first and second layers of this network consist of 300 and 100 nodes, respectively. The benchmark used for evaluating the results is the MNIST-Digit dataset \cite{deng2012mnist}, which is for handwriting digit recognition. Each sample of this dataset comes in a 28×28 pixels image related to one digit. We flatten each image into a 1D vector with $28\times28 =784$ parameters. The input size of the network was chosen equal to the input vector dimension, 784. The output layer has ten nodes, which represent ten classes, and each class is linked to one digit. The total number of weights is 266,610. MNIST-Digit has 60,000 images for training and 10,000 images for testing. Fig.~\ref{arch} illustrates the network architecture and the number of nodes for each layer. Activation functions for hidden layers one and two are ReLU \cite{agarap2018deep}, and for the output layer, it is Softmax. The loss function is categorical cross-entropy.

A simple ANN is still a very high-dimensional optimization problem. Gradient-based algorithms can train these networks with high accuracy, but non-gradient-based algorithms such as DE are remarkably time-consuming, and this makes them impractical for training such huge-scale networks, especially on a personal computer with only one CPU. On the other hand, we know that the CS algorithm shrinks the search space exponentially. Therefore, we have tailored it for training fully connected networks.  

\subsection{Metrics}
To assess the performance of training algorithms, we use accuracy, precision, and recall as metrics. Precision is the probability that an object is predicted correctly as a member of class $x$, given that it is returned by the system as a member of that class. Recall is the probability that an object of class $x$ is predicted correctly. 



\subsection{Numerical Results and Analysis}


\subsubsection{Comparing the Proposed Method with SGD}

In this section, the results of the two-extreme-point CS are compared with SGD, which is a well-known algorithm for training neural networks, when all or a small subset (e.g., 1/60) of the data is used for training. We want to see the effect of training on a small dataset. This is the case in many real-world applications because there is no access to a large labeled dataset most of the time.

\begin{figure}
\centering
\includegraphics[width=0.80\linewidth]{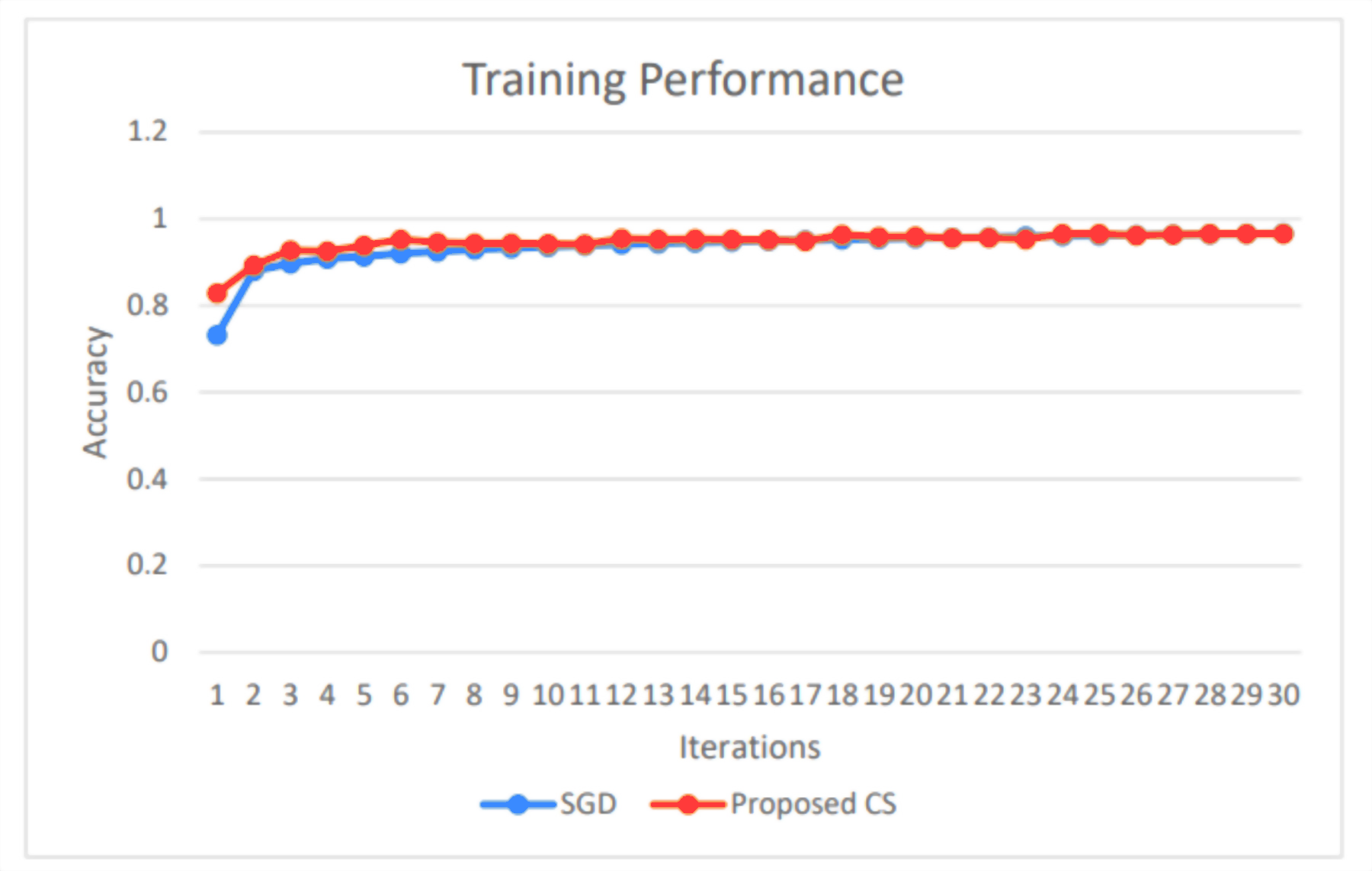}
\caption{Comparing the training performance plots of SGD and the proposed CS by feeding all training data}
\label{SGD-train-res-1}
\end{figure}

\begin{figure}
\centering
\includegraphics[width=0.80\linewidth]{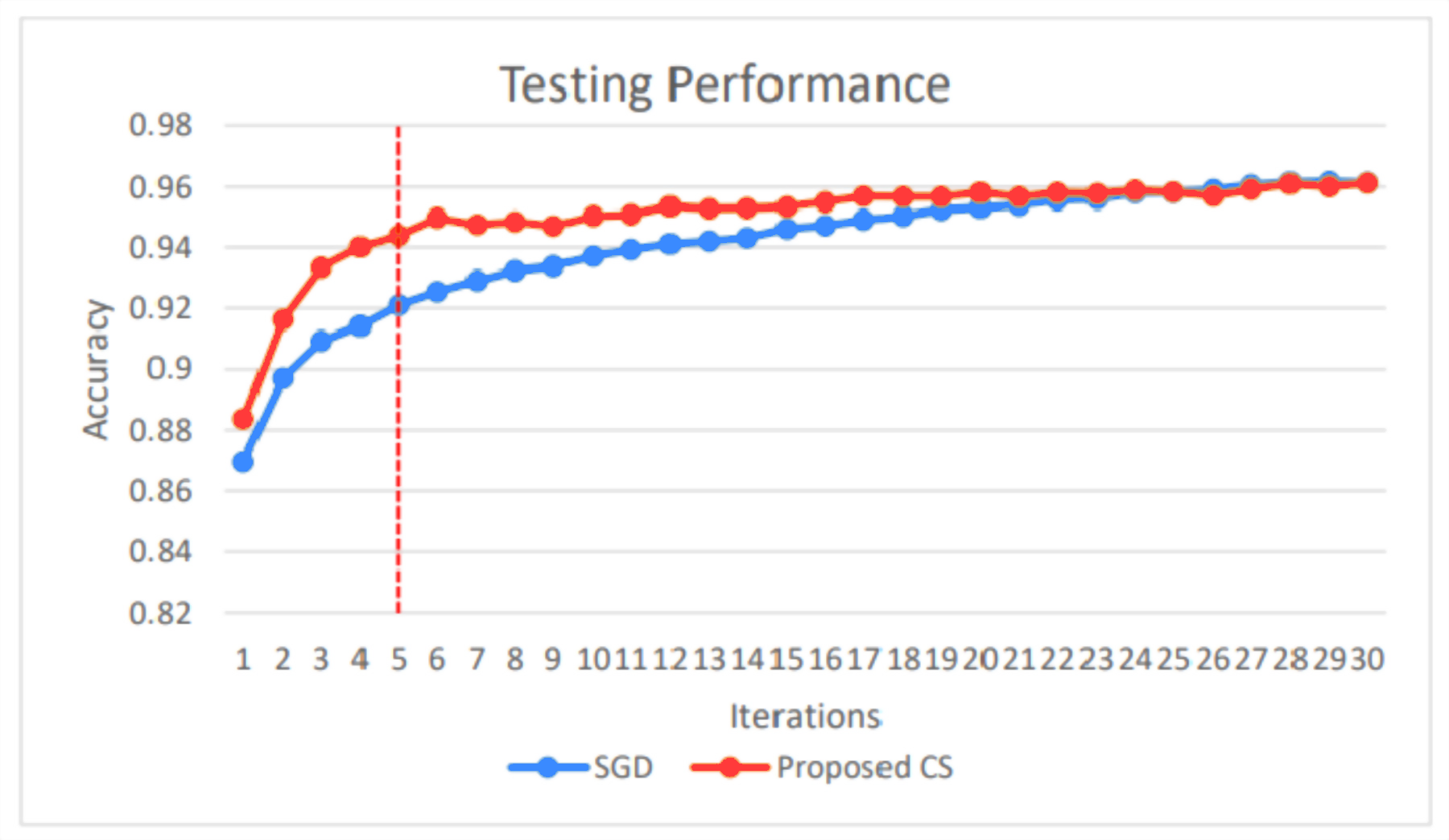}
\caption{Comparing the testing performance plots of SGD and the proposed CS by feeding all training data}
\label{SGD-test-res-1}
\end{figure}

First, we trained the network with 266,610 weights using all training samples from the MNIST dataset, including 60,000 instances. The performance plot of training and testing is provided in Figs.~\ref{SGD-train-res-1} and \ref{SGD-test-res-1}, respectively. The proposed CS starts with a better accuracy value and converges to the optimum point faster than SGD. However, in the end, the results show almost the same test accuracy for both methods. So, the suggested method can be regarded as an alternative for training neural networks. The proposed CS shows a remarkable ability to optimize very large-scale problems. Training and testing results are reported in Table~\ref{tab-SGD-1}.

\begin{table}[htbp]
\caption{Comparing the results of two-extreme-point CS and SGD when feeding all training data}
\begin{center}
\begin{tabular}{|c|c|c|}
\hline
               & SGD     & Proposed CS \\ \hline
Train Accuracy & 96.65\% & 96.59\%     \\ \hline
Test Accuracy  & 96.14\% & 96.12\%     \\ \hline
\end{tabular}
\label{tab-SGD-1}
\end{center}
\end{table}

In Fig.~\ref{SGD-test-res-1}, the dashed red line shows if the training process is stopped after five iterations, which is equal to 10D fitness calls, the accuracy is relatively high (92.11\%). This early stopping is very beneficial in evolving neural network processes. Assume that the hyperparameters of an ANN or its structure should be optimized as in \cite{rokhsatyazdi2020optimizing} (evolving deep neural networks), in which the network should be trained for each fitness call. Therefore, many fitness calls are needed for the optimization process, so the network should be trained many times which is extremely time-consuming. If early stopping leads to high accuracy, which is the case with our proposed CS method, optimization can be accelerated significantly. 

To continue, 1000 training samples from the MNIST Fashion dataset, which includes 60,000 training samples in total, are selected for training the neural network. This reflects the lack of enough labeled training data and is one of the main challenges in optimizing neural networks. Table~\ref{tab-SGD-2} reports the results for three test cases: SGD, the proposed CS, and the hybrid scheme. In the hybrid method, the network is trained using CS only for one iteration. In the next iterations (epochs), the adjusted weights by CS are passed to SGD as initial points. SGD optimizes the weights in the following epochs. The hybrid method benefits from both the advantages of CS, which is higher accuracy, and SGD, which is higher speed. From the table, CS achieves the best accuracy while SGD cannot reach notable accuracy with small-size training data. The hybrid method outperforms SGD in terms of training and test accuracies. Although it could not reach CS accuracy, it is a faster technique than CS.

\begin{table}[htbp]
\caption{Comparing SGD, two-extreme-point CS, and hybrid optimizer by feeding a small subset of training data}
\begin{center}
\begin{tabular}{|c|c|c|c|}
\hline
               & SGD    & Proposed CS     & Hybrid \\ \hline
Train Accuracy & 75.2\% & \textbf{96.9\%} & 86.2\% \\ \hline
Test Accuracy  & 71.1\% & \textbf{77.4\%} & 75.7\% \\ \hline
\end{tabular}
\label{tab-SGD-2}
\end{center}
\end{table}

\begin{figure}
\centering
\includegraphics[width=0.80\linewidth]{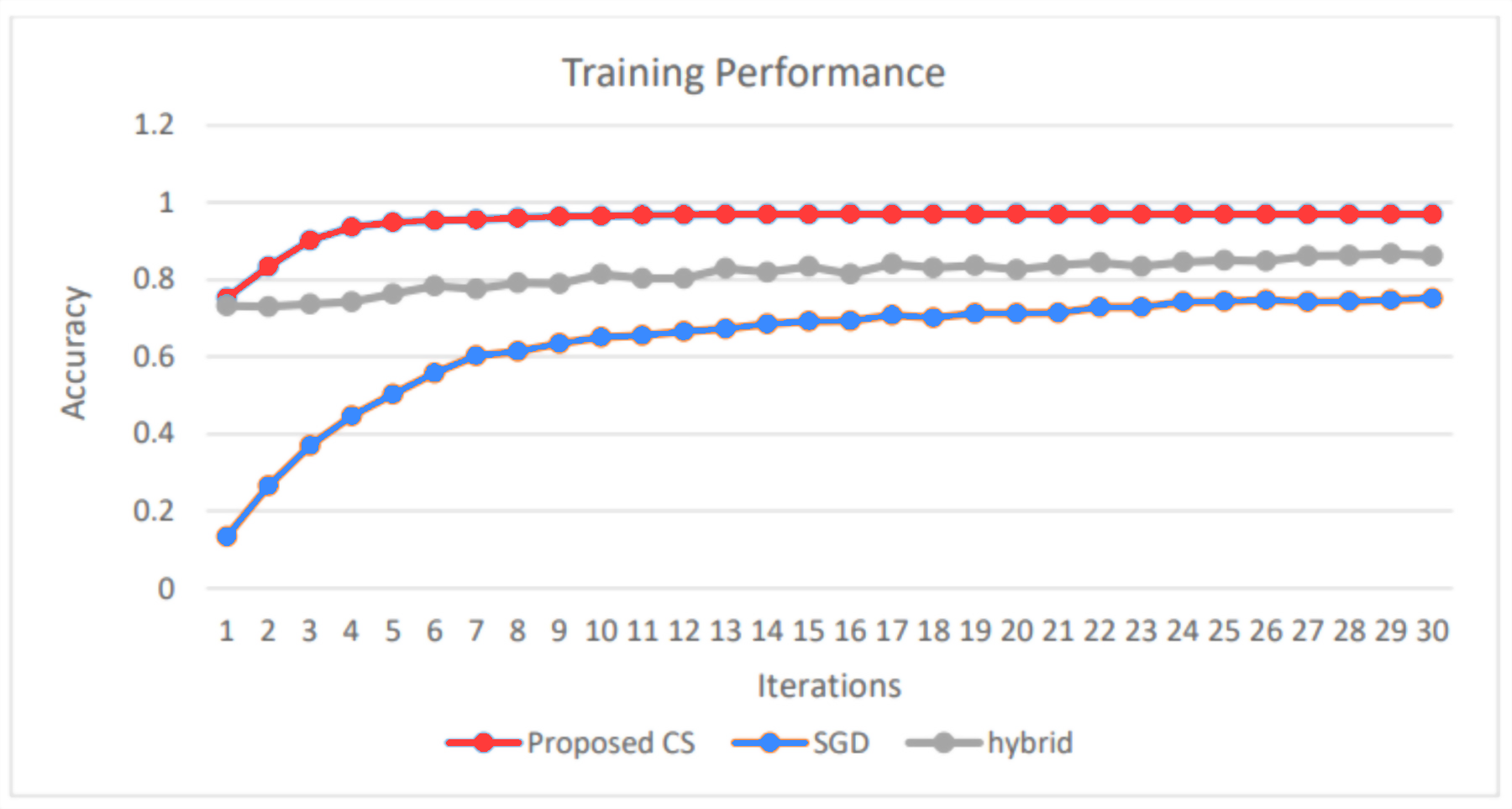}
\caption{Comparing the training performance plots of SGD, the proposed CS, and their hybrid by feeding a small subset (1/60) of training data}
\label{SGD-train-res-2}
\end{figure}

\begin{figure}
\centering
\includegraphics[width=0.80\linewidth]{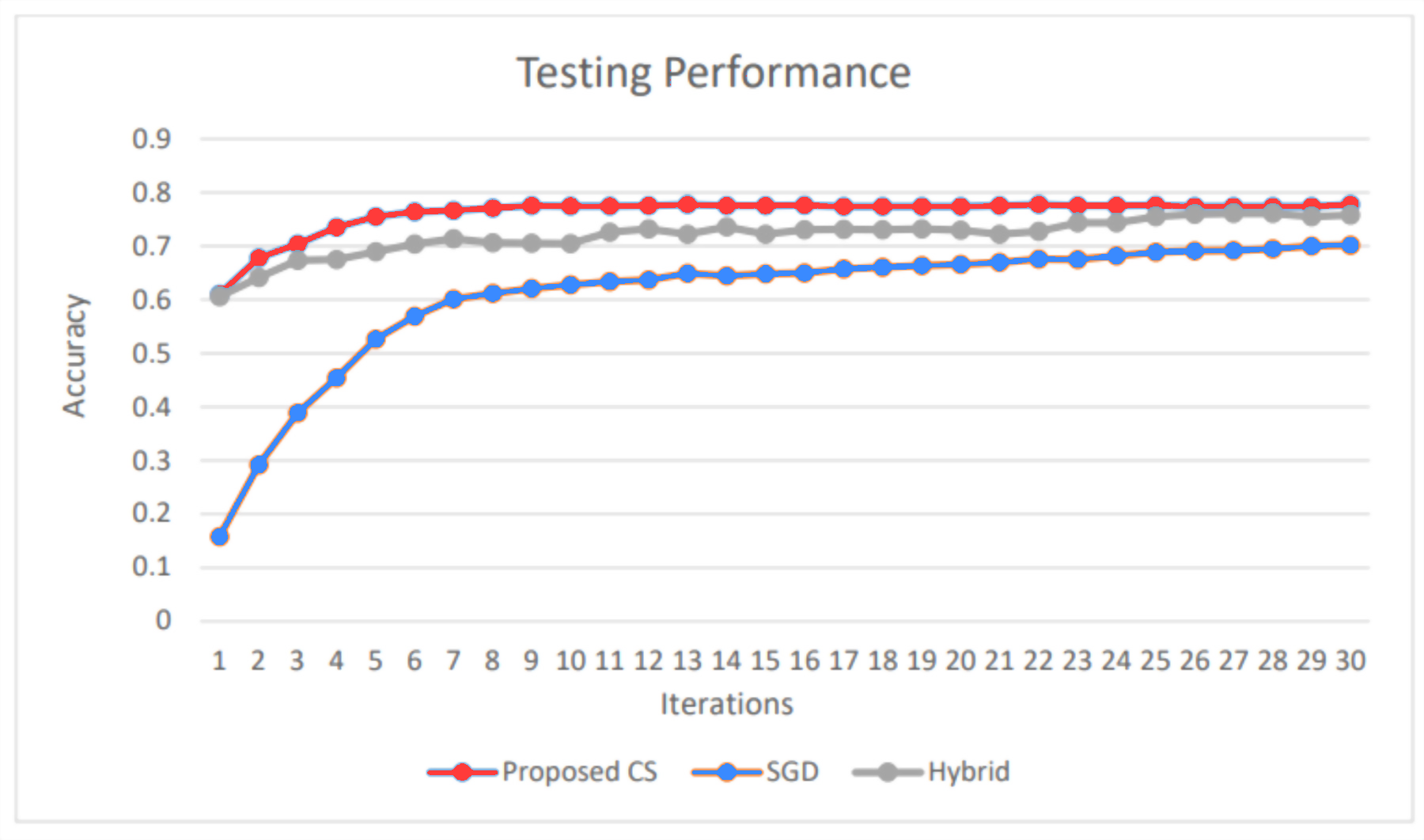}
\caption{Comparing the testing performance plots of SGD, the proposed CS, and their hybrid by feeding a small subset (1/60) of training data}
\label{SGD-test-res-2}
\end{figure}

Performance plots for training and testing of this series of experiments are illustrated in Figs.~\ref{SGD-train-res-2} and ~\ref{SGD-test-res-2}, respectively. As it is illustrated, the proposed CS can find a better set of weights with a small number of iterations. This fast convergence rate is very beneficial since optimizing weights in neural networks is demanding. So, a fast convergence rate reduces training time.
Meanwhile, we should note that, by using a smaller set of training samples, CS would perform even faster while achieving better accuracy than SGD.

\begin{figure*}
\centering
\begin{tabular}{cc}
\includegraphics[width=0.30\linewidth]{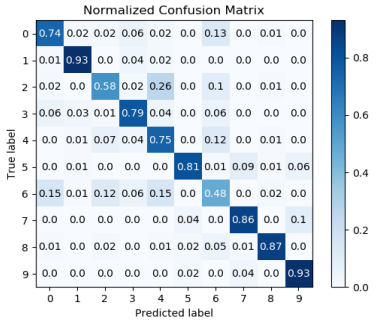} &\includegraphics[width=0.30\linewidth]{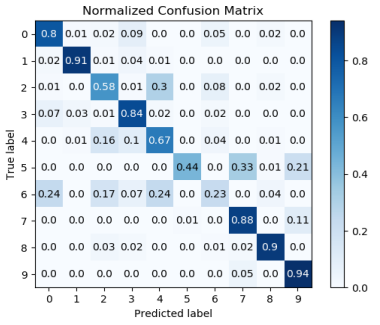}\\
(a) Proposed CS & (b) SGD\\
\end{tabular}
\caption{Comparing confusion matrices for the proposed method and SGD on MNIST
}
\label{CM}
\end{figure*}

To better understand network performance, the confusion matrix, precision, and recall are also reported besides accuracy. The normalized confusion matrix for SGD and the proposed CS is shown in Fig.~\ref{CM}. For the proposed CS, the number and amount of confusion between classes are less than SGD, which indicates superior classification ability. In addition, Table~\ref{tab-prec-recall} reports the average precision and recall over all classes. For both precision and recall cases, CS achieves significantly better results than SGD. It proves our method's superiority over SGD in classification.

\begin{table}[htbp]
\caption{Comparing average precision and recall over all classes}
\begin{center}
\begin{tabular}{|c|c|c|}
\hline
            & Average Precision & Average Recall  \\ \hline
SGD         & 0.7259            & 0.7185          \\ \hline
Proposed CS & \textbf{0.7774}   & \textbf{0.7743} \\ \hline
\end{tabular}
\label{tab-prec-recall}
\end{center}
\end{table}

\subsubsection{Effect of Initialization}
Table~\ref{tab-init} shows the test accuracy of the network when using different initialization approaches. The center initialization which sets all weights to zero as the initial values reached the accuracy of 93.71\%. For uniform random initialization, the results for different intervals are reported. In addition,  different values for mean and standard deviation for random initialization with a normal distribution are also evaluated. In conclusion, random normal initialization ($\mu=0, \sigma=0.1$) results in the highest accuracy.

\begin{table*}[htbp]
\caption{Comparing the effect of different initialization approaches on the test accuracy}
\begin{center}
\begin{tabular}{|c|cccccc|}
\hline
Fixed Initialization & \multicolumn{6}{c|}{Random Initialization}                                                                                                                                                                      \\ \hline
Center               & \multicolumn{2}{c|}{Uniform}                                              & \multicolumn{4}{c|}{Center Normal Distribution}                                                                                     \\ \hline
Value = 0            & \multicolumn{1}{c|}{$R \in [-2,2]$} & \multicolumn{1}{c|}{$R \in [-1,1]$} & \multicolumn{1}{c|}{$\mu = 0, \sigma = 0.05$} & \multicolumn{1}{c|}{$\mu = 0, \sigma = 0.1$}   & \multicolumn{1}{c|}{$\mu = 0, \sigma = 0.2$} & $\mu = 0, \sigma = 0.3$ \\ \hline
93.71\%              & \multicolumn{1}{c|}{91.2\%}         & \multicolumn{1}{c|}{92.63\%}        & \multicolumn{1}{c|}{93.64\%}         & \multicolumn{1}{c|}{\textbf{93.82\%}} & \multicolumn{1}{c|}{92.52\%}        & 92.3\%         \\ \hline
\end{tabular}
\label{tab-init}
\end{center}
\end{table*}

\subsubsection{Effect of Data Feeding}
\begin{figure}
\centering
\begin{tabular}{c}
\includegraphics[width=0.70\linewidth]{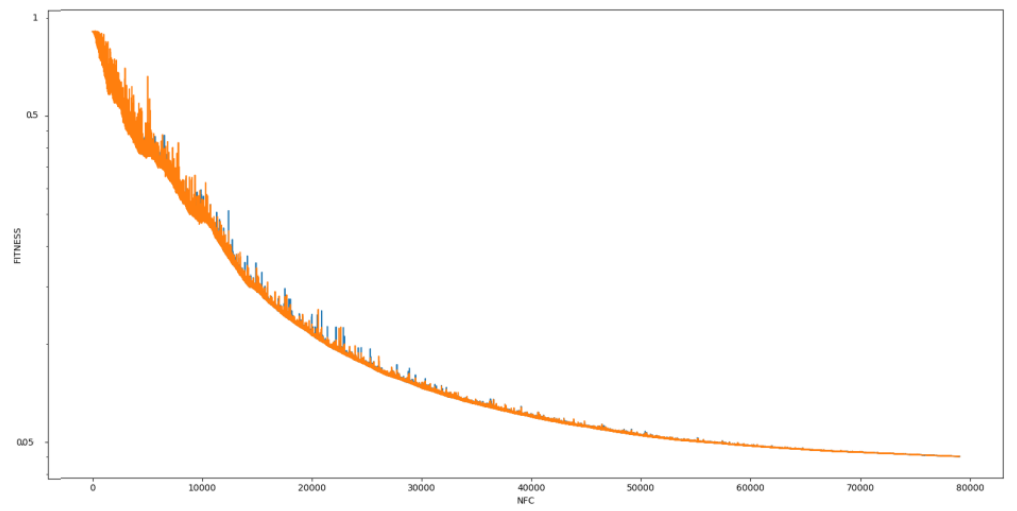} \\
(a) Feeding whole data in one batch \\
\includegraphics[width=0.70\linewidth]{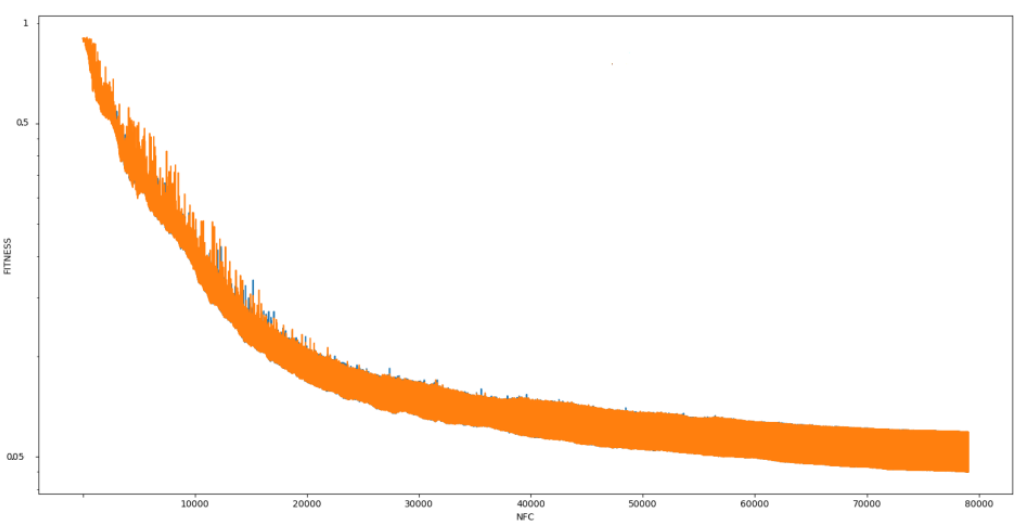} \\
(b) Feeding with separate folds (6 batches) \\
\includegraphics[width=0.70\linewidth]{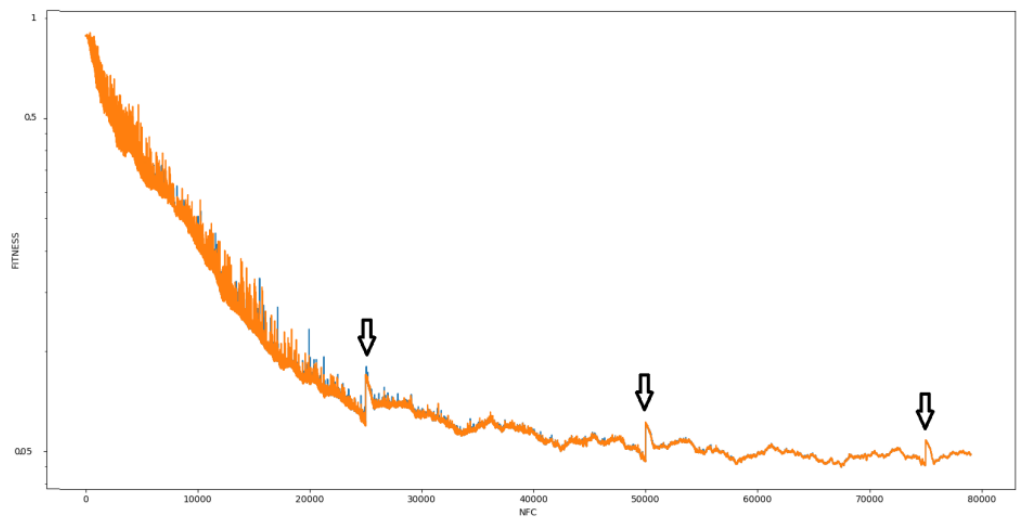} \\
(c) Feeding with sliding fold \\
\end{tabular}
\caption{Results of training the fully connected neural network with the proposed CS using various training data feeding methods}
\label{feed-res-fig}
\end{figure}

In this section, different ways of feeding training data into the network are investigated and the results are presented in Fig.~\ref{feed-res-fig}. Feeding with separate folds causes more fluctuations in training accuracy (Fig.~\ref{feed-res-fig}(b)). However, as shown in Fig.~\ref{feed-res-fig}(c), when it reaches the end of the training procedure of each fold, a spike appears in the performance plot. The sliding amount in this test is 100, i.e., each fold slides 100 samples each time. Table~\ref{tab-feed} shows the effect of different data feeding techniques on accuracy. The highest accuracy can be achieved by feeding the whole data in each function call, but the fastest training can be obtained using separate and sliding folds. With the same training speed, separate folds lead to the better accuracy.

The number of folds for feeding training data is a parameter that can be studied. Higher numbers even make the optimization faster, but maybe a trade-off is needed because the smaller size of folds may degrade accuracy of training.

\begin{table*}[htbp]
\caption{Accuracy resulted from different feeding data approaches}
\begin{center}
\begin{tabular}{|c|c|c|c|}
\hline
                                                                                  & Feed whole data in one batch & Feed with separate folds & \begin{tabular}[c]{@{}c@{}}Feed with sliding fold\\ (sliding amount = 100)\end{tabular} \\ \hline
Fixed center initialization                                                       & 94.66\%                      & 93.71\%                  & 93.52\%                                                                                 \\ \hline
\begin{tabular}[c]{@{}c@{}}Random center initialization\\ $\mu=0, \sigma=0.1$\end{tabular} & \textbf{94.74\%}             & 93.82\%                  & 93.70\%                                                                                 \\ \hline
Speed-up                                                                          & 1x                           &$\sim$ 6x                       &$\sim$ 6x                                                                                      \\ \hline
\end{tabular}
\label{tab-feed}
\end{center}
\end{table*}

\subsubsection{Effect of Different Bundle Sizes (BS)}
As mentioned earlier, bundling weights is a sort of dimension reduction. To explore the effect of bundle size (BS), some experiments were conducted. It is noticeable that larger bundle sizes lead to a lower number of function calls (NFC), and vice versa. For a fair comparison, the NFC for all tests is considered a fixed number. So, the number of iterations varies based on BS. When BS is 100, the number of iterations is 40, and when BS is 25, the number of iterations is decreased to 10 to keep the same NFC. BS can be considered fixed for all iterations  or it may adaptively change in different iterations. For example, in one of our tests, BS is set to 25 for the first five iterations (half of the total NFC) and changed to 100 for the next 20 iterations. An immature convergence may occur when there are fewer iterations. Table~\ref{tab-BS} reports the results of three test scenarios: 
In the first scenario, we set the BS to 100 for 40 iterations. In the second one, we run 20 iterations with BS=100 and change it to 25 for the next 5 iterations. In the last scenario, BS is increased from 25 to 100 after 5 iterations.

\begin{table*}[htbp]
\caption{The effect of bundle size, explored in three scenarios: In the first scenario, BS is set to 100 for 40 iterations. In the second one, we run 20 iterations with BS=100 and change it to 25 for the next 5 iterations. In the last scenario, BS is increased from 25 to 100 after 5 iterations.}
\begin{center}
\begin{tabular}{|c|c|c|c|}
\hline
Bundle Size (BS)             & \includegraphics[width=0.13\linewidth]{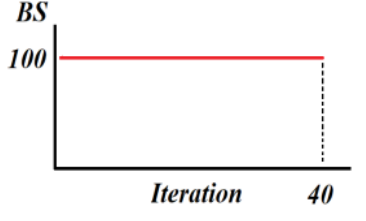} & \includegraphics[width=0.12\linewidth]{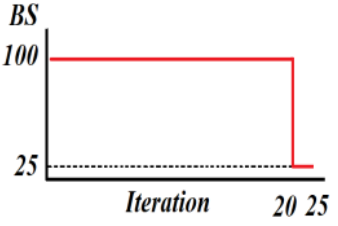} & \includegraphics[width=0.13\linewidth]{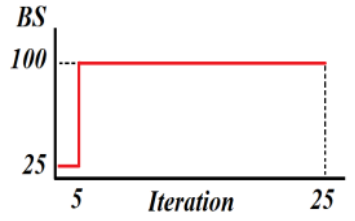} \\ \hline
Feed with separate folds     & 93.82\%                      & 93.95\%                      & 95.23\%                      \\ \hline
Feed whole data in one batch & 94.74\%                      & 94.88\%                      & \textbf{96.12\%}             \\ \hline
\end{tabular}
\label{tab-BS}
\end{center}
\end{table*}

Results show that when BS is low (i.e., 25) in the first 5 iterations, accuracy is higher than other cases. Due to the lower BS in the first iterations, this scenario indicates better exploration and higher accuracy. From the 6th iteration, we increase the BS to 100 for the remaining 20 iterations to accelerate the optimization process.

\section{CONCLUSION REMARKS}
In this paper, a novel method for training a neural network  was presented as a high-dimensional optimization problem, a method based on two-extreme-point bundling CS. The proposed method is a simple gradient-free algorithm where a group of variables (weights) are bundled together to be optimized simultaneously. For each bundle, the extreme points of box-constraints are evaluated to determine the optimal values for all variables in the bundle. 

We explored the effect of different initialization methods, bundle sizes, and loading data approaches to present the most effective method. Based on the obtained results, random normal initialization, feeding whole training data in one batch, and using lower bundle sizes in first iterations lead to better accuracy. However, combining CS with SGD, feeding data in separate folds, and increasing the bundle size accelerate training.

 Our suggested CS algorithm can be a feasible alternative method to train artificial neural networks. It can find good solutions with very small NFC, which can be very beneficial in expensive optimization processes, where each function call incurs high costs, as well as in evolving neural network procedures aimed at acceleration. This algorithm treats the problem as a black-box optimization task, so it can train complex network structures such as graph-based networks, which is not possible with gradient-based techniques. Moreover, this algorithm can be used with non-differentiable activation functions and also tailored to optimize networks for multi-objective/multi-loss problems. In addition, the experiments showed that the lack of enough labeled training data, which is the case in many real-world applications, is in favor of our method. Future works should conduct experiments on various sizes and architectures of deep neural network. 

\bibliography{ref}

\begin{thebibliography}{10}
\providecommand{\url}[1]{#1}
\csname url@samestyle\endcsname
\providecommand{\newblock}{\relax}
\providecommand{\bibinfo}[2]{#2}
\providecommand{\BIBentrySTDinterwordspacing}{\spaceskip=0pt\relax}
\providecommand{\BIBentryALTinterwordstretchfactor}{4}
\providecommand{\BIBentryALTinterwordspacing}{\spaceskip=\fontdimen2\font plus
\BIBentryALTinterwordstretchfactor\fontdimen3\font minus \fontdimen4\font\relax}
\providecommand{\BIBforeignlanguage}[2]{{%
\expandafter\ifx\csname l@#1\endcsname\relax
\typeout{** WARNING: IEEEtran.bst: No hyphenation pattern has been}%
\typeout{** loaded for the language `#1'. Using the pattern for}%
\typeout{** the default language instead.}%
\else
\language=\csname l@#1\endcsname
\fi
#2}}
\providecommand{\BIBdecl}{\relax}
\BIBdecl

\bibitem{kaveh2022application}
M.~Kaveh and M.~S. Mesgari, ``Application of meta-heuristic algorithms for training neural networks and deep learning architectures: A comprehensive review,'' \emph{Neural Processing Letters}, pp. 1--104, 2022.

\bibitem{javanshir2023training}
A.~Javanshir, T.~T. Nguyen, M.~P. Mahmud, and A.~Z. Kouzani, ``Training spiking neural networks with metaheuristic algorithms,'' \emph{Applied Sciences}, vol.~13, no.~8, p. 4809, 2023.

\bibitem{leung2003tuning}
F.~H.-F. Leung, H.-K. Lam, S.-H. Ling, and P.~K.-S. Tam, ``Tuning of the structure and parameters of a neural network using an improved genetic algorithm,'' \emph{IEEE Transactions on Neural networks}, vol.~14, no.~1, pp. 79--88, 2003.

\bibitem{meissner2006optimized}
M.~Meissner, M.~Schmuker, and G.~Schneider, ``Optimized particle swarm optimization (opso) and its application to artificial neural network training,'' \emph{BMC bioinformatics}, vol.~7, no.~1, pp. 1--11, 2006.

\bibitem{geethanjali2008pso}
M.~Geethanjali, S.~M.~R. Slochanal, and R.~Bhavani, ``Pso trained ann-based differential protection scheme for power transformers,'' \emph{Neurocomputing}, vol.~71, no. 4-6, pp. 904--918, 2008.

\bibitem{yu2008evolving}
J.~Yu, S.~Wang, and L.~Xi, ``Evolving artificial neural networks using an improved pso and dpso,'' \emph{Neurocomputing}, vol.~71, no. 4-6, pp. 1054--1060, 2008.

\bibitem{zhou2006pso}
J.~Zhou, Z.~Duan, Y.~Li, J.~Deng, and D.~Yu, ``Pso-based neural network optimization and its utilization in a boring machine,'' \emph{Journal of Materials Processing Technology}, vol. 178, no. 1-3, pp. 19--23, 2006.

\bibitem{carvalho2007particle}
M.~Carvalho and T.~B. Ludermir, ``Particle swarm optimization of neural network architectures andweights,'' in \emph{7th International Conference on Hybrid Intelligent Systems (HIS 2007)}.\hskip 1em plus 0.5em minus 0.4em\relax IEEE, 2007, pp. 336--339.

\bibitem{grimaldi2004pso}
E.~A. Grimaldi, F.~Grimaccia, M.~Mussetta, and R.~Zich, ``Pso as an effective learning algorithm for neural network applications,'' in \emph{Proceedings. ICCEA 2004. 2004 3rd International Conference on Computational Electromagnetics and Its Applications, 2004.}\hskip 1em plus 0.5em minus 0.4em\relax IEEE, 2004, pp. 557--560.

\bibitem{ilonen2003differential}
J.~Ilonen, J.-K. Kamarainen, and J.~Lampinen, ``Differential evolution training algorithm for feed-forward neural networks,'' \emph{Neural Processing Letters}, vol.~17, pp. 93--105, 2003.

\bibitem{jameson1995gradient}
A.~Jameson, ``Gradient based optimization methods,'' \emph{MAE Technical Report No}, no. 2057, 1995.

\bibitem{kingma2014adam}
D.~P. Kingma and J.~Ba, ``Adam: A method for stochastic optimization,'' \emph{arXiv preprint arXiv:1412.6980}, 2014.

\bibitem{schwefel1993evolution}
H.-P.~P. Schwefel, \emph{Evolution and optimum seeking: the sixth generation}.\hskip 1em plus 0.5em minus 0.4em\relax John Wiley \& Sons, Inc., 1993.

\bibitem{bogani2009generalized}
C.~Bogani, M.~Gasparo, and A.~Papini, ``Generalized pattern search methods for a class of nonsmooth optimization problems with structure,'' \emph{Journal of Computational and Applied Mathematics}, vol. 229, no.~1, pp. 283--293, 2009.

\bibitem{tzinis2019bootstrapped}
E.~Tzinis, ``Bootstrapped coordinate search for multidimensional scaling,'' \emph{arXiv preprint arXiv:1902.01482}, 2019.

\bibitem{bidgoli2021memetic}
A.~A. Bidgoli and S.~Rahnamayan, ``Memetic differential evolution using coordinate descent,'' in \emph{2021 IEEE Congress on Evolutionary Computation (CEC)}.\hskip 1em plus 0.5em minus 0.4em\relax IEEE, 2021, pp. 359--366.

\bibitem{frandi2014coordinate}
E.~Frandi and A.~Papini, ``Coordinate search algorithms in multilevel optimization,'' \emph{Optimization Methods and Software}, vol.~29, no.~5, pp. 1020--1041, 2014.

\bibitem{tseng2001convergence}
P.~Tseng, ``Convergence of a block coordinate descent method for nondifferentiable minimization,'' \emph{Journal of optimization theory and applications}, vol. 109, no.~3, pp. 475--494, 2001.

\bibitem{nikbakht2023multi}
F.~Nikbakhtsarvestani, A.~Asilian~Bidgoli, M.~Ebrahimi, and S.~Rahnamayan, ``Multi-objective coordinate search optimization,'' pp. 1--7, 2023.

\bibitem{rahnamayan2009toward}
S.~Rahnamayan and G.~G. Wang, ``Toward effective initialization for large-scale search spaces,'' \emph{Trans Syst}, vol.~8, no.~3, pp. 355--367, 2009.

\bibitem{mahdavi2016center}
S.~Mahdavi, S.~Rahnamayan, and K.~Deb, ``Center-based initialization of cooperative co-evolutionary algorithm for large-scale optimization,'' in \emph{2016 IEEE Congress on Evolutionary Computation (CEC)}.\hskip 1em plus 0.5em minus 0.4em\relax IEEE, 2016, pp. 3557--3565.

\bibitem{rahnamayan2009center}
S.~Rahnamayan and G.~G. Wang, ``Center-based initialization for large-scale black-box problems,'' in \emph{Proceedings of the 8th WSEAS international conference on Artificial intelligence, knowledge engineering and data bases}, 2009, pp. 531--541.

\bibitem{masters2018revisiting}
D.~Masters and C.~Luschi, ``Revisiting small batch training for deep neural networks,'' \emph{arXiv preprint arXiv:1804.07612}, 2018.

\bibitem{lu2019not}
Y.~Lu and R.~Yang, ``Not all features are equal: Feature leveling deep neural networks for better interpretation,'' \emph{arXiv preprint arXiv:1905.10009}, 2019.

\bibitem{deng2012mnist}
L.~Deng, ``The mnist database of handwritten digit images for machine learning research,'' \emph{IEEE Signal Processing Magazine}, vol.~29, no.~6, pp. 141--142, 2012.

\bibitem{agarap2018deep}
A.~F. Agarap, ``Deep learning using rectified linear units (relu),'' \emph{arXiv preprint arXiv:1803.08375}, 2018.

\bibitem{rokhsatyazdi2020optimizing}
E.~Rokhsatyazdi, S.~Rahnamayan, H.~Amirinia, and S.~Ahmed, ``Optimizing lstm based network for forecasting stock market,'' in \emph{2020 IEEE congress on evolutionary computation (CEC)}.\hskip 1em plus 0.5em minus 0.4em\relax IEEE, 2020, pp. 1--7.

\end{thebibliography}
\bibliographystyle{IEEEtran}

\vspace{12pt}

\end{document}